\documentclass[letterpaper, 10 pt, conference]{ieeeconf} 
\IEEEoverridecommandlockouts
\usepackage{graphicx}
\usepackage{mathptmx}
\usepackage{times}
\usepackage[stretch=40,shrink=40]{microtype}
\usepackage{amsmath}
\usepackage{amssymb}
\usepackage{booktabs}
\usepackage{colortbl}
\usepackage{multirow}
\usepackage{rotating}
\usepackage{tikz}
\usetikzlibrary{positioning,fit,calc}
\usepackage{capt-of}
\usepackage{placeins}
\usepackage{stfloats}
\usepackage{algorithm}
\usepackage{algpseudocode}
\usepackage{eso-pic}
\newcommand{\IEEEsubmittednotice}{%
  \AddToShipoutPictureBG*{%
    \AtPageLowerLeft{%
      \raisebox{8mm}{\makebox[\paperwidth]{%
        \fbox{\parbox[b]{0.5\paperwidth}{\footnotesize
          This work has been submitted to the IEEE for possible publication.          
          Copyright may be transferred without notice, after which this
          version may no longer be accessible.}}}}}}}

\makeatletter
\let\NAT@parse\undefined
\makeatother
\usepackage[hidelinks]{hyperref}

\newif\iffigplaceholders
\figplaceholdersfalse
\newcommand{\figplaceholder}[3]{%
  \begingroup
    \setlength{\fboxsep}{0pt}%
    \fbox{\begin{minipage}[c][#2][c]{\dimexpr#1-2\fboxrule\relax}%
      \centering\footnotesize\baselineskip=10pt #3%
    \end{minipage}}%
  \endgroup}
\newcommand{\figslot}[4]{%
  \iffigplaceholders
    \figplaceholder{#1}{#2}{#4}%
  \else
    \IfFileExists{#3}{\includegraphics[width=#1]{#3}}%
      {\figplaceholder{#1}{#2}{#4}}%
  \fi}

\newlength{\tablenotetopsep}
\newlength{\tablenotebodypullup}
\newcommand{\tablenotebefore}{\par\vspace{\tablenotetopsep}\noindent}
\newcommand{\tablenoteafter}{\par\vspace{-\tablenotebodypullup}}

\newenvironment{papertablenote}{%
  \tablenotebefore
  \begin{minipage}{0.98\linewidth}\scriptsize
  \setlength{\parindent}{0pt}%
}{\end{minipage}\tablenoteafter}

\title{\LARGE \bf
AnchorHEC: Targetless Hand-Eye Calibration\\ of Wrist and Fixed Cameras
}

\author{
Guebin Hwang, Taeyeop Lee, Yeonseo Lee, Hyosup Shin,
and Sungho Jo$^{\dagger}$%
\thanks{All authors are with the
Korea Advanced Institute of Science and Technology (KAIST), Daejeon, South Korea.}%
\thanks{$^{\dagger}$Corresponding author}%
}

\begin{document}
\bstctlcite{IEEEtranBSTCTL:etal}

\maketitle
\IEEEsubmittednotice
\thispagestyle{empty}
\pagestyle{empty}

\begin{abstract}
Recent robot systems combine wrist and fixed cameras, requiring
camera-to-robot transforms for both. Most hand-eye calibration methods
rely on targets or robot-body observations. Scene-based
methods using 3D reconstruction require neither but address only wrist
cameras. We introduce AnchorHEC, a targetless hand-eye calibration
framework for wrist and fixed cameras in a single procedure using RGB
images, forward kinematics (FK), and known camera intrinsics.
We initialize both camera types using FK and a shared 3D foundation model
reconstruction. FK-constrained wrist-image bundle adjustment reduces the
impact of reconstructed pose and scale errors. If this refinement fails
or retains too few inliers, we relax the constraints on potentially
inaccurate FK poses and lens parameters. We use the refined wrist
transform and FK to triangulate metric points and estimate fixed-camera
poses.
Across three backbones, calibration errors in simulation decrease by
86.4\% in translation and 71.6\% in rotation, averaged over both camera
types, relative to reconstruction-based baselines. We also validate
AnchorHEC on a physical robot.
Project page: \url{https://hwanggb.github.io/AnchorHEC/}
\end{abstract}

\section{INTRODUCTION}
\label{sec:intro}

Camera-to-robot transforms enable vision-based control and
policies that predict robot actions in camera
coordinates~\cite{OC-VLA} or explicitly condition on camera
pose~\cite{KnowYourCamera}.
Recent manipulation systems combine wrist-mounted
(eye-in-hand) cameras for close-up views with fixed
(eye-to-hand) cameras for broader workspace
coverage~\cite{DROID,ALOHA}.
Using both views in robot coordinates requires the rigid transforms
from the wrist camera to the end effector and from each fixed camera
to the robot base, which hand-eye calibration estimates~\cite{Tsai1989,ATOM2022}.
\looseness=-1

Existing hand-eye calibration methods fall into three categories 
by what they observe: target-based, robot-body-based, and scene-based. 
The first two handle both camera types. 
Target-based methods~\cite{ATOM2022,Pedrosa2021,Zhou2024TIE} use known
calibration patterns, such as ArUco~\cite{ArUco} or AprilTag~\cite{AprilTag}.
Camera remounting, changes in fixed-camera pose relative to the robot
base~\cite{Miseikis2016}, and vibration-induced mounting
drift~\cite{Kessens2020RoMan} can require recalibration.
Each recalibration requires manually setting up a target board 
and keeping it within the camera's field of view throughout data collection.
\looseness=-1

\begin{figure}[!t]
\centering
\figslot{\columnwidth}{52mm}{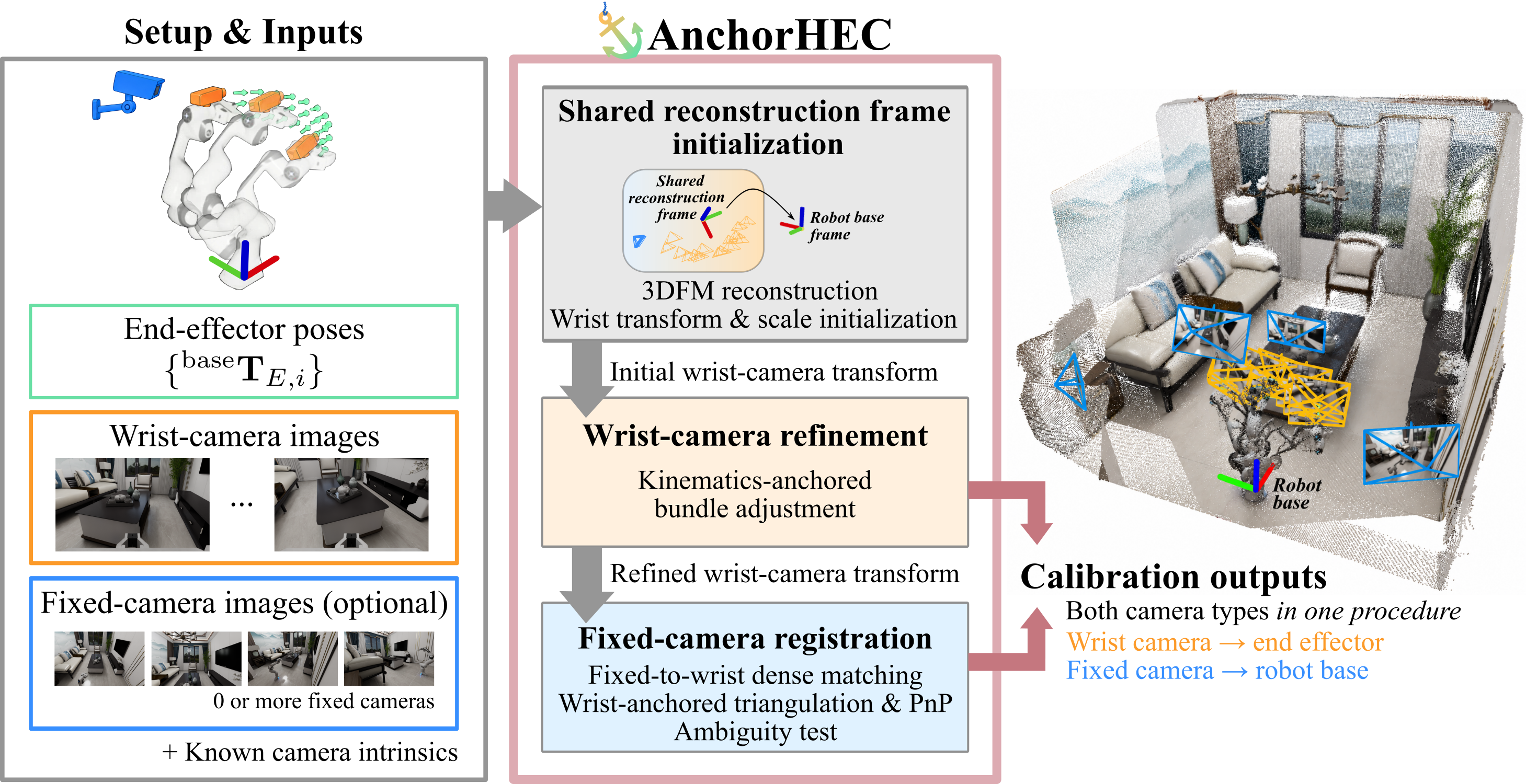}{%
  F1 CONCEPT OVERVIEW, USER ARTWORK SLOT\\[3pt]
  LEFT: moving wrist + fixed cameras observing one scene\\
  CENTRE: shared 3D reconstruction + FK-constrained wrist refinement\\
  RIGHT: cameras registered to the robot base\\[3pt]
  Inputs: scene RGB, robot poses, known intrinsics\\
  Outputs: camera-to-robot transforms for wrist and fixed cameras}
\caption{AnchorHEC overview. Joint reconstruction and forward kinematics
(FK) initialize camera-to-robot transforms for wrist and fixed cameras. 
The refined wrist transform and FK anchor metric scene geometry 
for fixed-camera registration. Green markers denote end-effector poses. 
Orange and blue denote wrist and fixed cameras, respectively.}
\label{fig:overview}
\end{figure}

Robot-body-based methods avoid calibration targets by estimating camera-to-robot
transforms from observations of the robot itself. Some methods require
robot-specific visual training~\cite{Lee2020,Valassakis2021}. CAD-based
approaches use robot or end-effector models~\cite{Wu2025,EasyHeC,EasyHeCpp},
while reference-point tracking requires selecting a suitable point with known
coordinates on the robot~\cite{Kalib}. 
These methods therefore depend on models or reference points 
prepared for each robot or end effector, and they cannot use images 
in which the required robot features are occluded or outside the field of view.
\looseness=-1

Scene-based methods remove the need for calibration targets and robot-body
observations by estimating camera motion through 3D reconstruction of the
surrounding scene~\cite{Andreff2001,Heller2011,JCR,BiJCR,Calib3R}.
Existing scene-based methods, however, handle only wrist cameras.
They recover the wrist transform by relating camera motion between views
to end-effector motion between those views. Fixed cameras do not move with
the end effector, so calibrating both camera types requires an additional step to
obtain the fixed-camera extrinsics.
\looseness=-1

Recent scene-based methods use 3D foundation models
(3DFMs)~\cite{JCR,BiJCR,Calib3R}.
These models estimate camera poses and scene geometry in a common
coordinate frame from images captured at widely separated
viewpoints~\cite{DUSt3R,MASt3R,VGGT}.
Using such reconstructions to calibrate wrist and fixed cameras in a single
procedure remains underexplored.

The key observation for extending scene-based calibration 
to both camera types is that fixed-camera poses can be related 
to those of the moving wrist camera through a joint reconstruction of their images.
However, recovering camera-to-robot transforms from this reconstruction still
requires metric scale and alignment with the robot base.
Moreover, errors in the reconstructed camera poses can propagate
to the calibration estimates.
\looseness=-1

To address these challenges, we present \textbf{AnchorHEC} (Fig.~\ref{fig:overview}),
a targetless hand-eye calibration framework for wrist and fixed cameras
in a single procedure without robot-body references (Table~\ref{tab:related_compare}).
The framework comprises three stages: shared reconstruction
frame initialization, wrist-camera refinement, and fixed-camera registration.
Wrist motion and forward kinematics (FK) constrain the metric scale and
robot-base alignment of the joint reconstruction, initializing
camera-to-robot transforms for both camera types.
\looseness=-1

\begin{table}[t]
\caption{Targetless Camera-to-Robot Calibration Methods}
\label{tab:related_compare}
\centering
\scriptsize
\setlength{\tabcolsep}{2.5pt}
\renewcommand{\arraystretch}{1.02}
\begin{tabular}{@{}
  >{\raggedright\arraybackslash}p{2.0cm}
  >{\centering\arraybackslash}p{0.72cm}
  >{\centering\arraybackslash}p{0.72cm}
  >{\centering\arraybackslash}p{0.72cm}
  >{\raggedright\arraybackslash}p{\dimexpr\columnwidth-4.33cm-8\tabcolsep\relax}@{}}
\toprule
Method & \shortstack{Wrist\\camera} & \shortstack{Fixed\\camera}
  & \shortstack{Single\\run} & \shortstack[l]{Robot-specific\\visual requirements} \\
\midrule
\rowcolor{black!6}
\multicolumn{5}{@{}l}{\textit{Robot-body-based}} \\
DREAM~\cite{Lee2020} & $\times$ & \checkmark & $\times$ & Robot-specific trained model \\
Valassakis~\cite{Valassakis2021} & \checkmark & $\times$ & $\times$ & Robot-specific trained model \\
FEEPE~\cite{Wu2025} & $\times$ & \checkmark & $\times$ & End-effector CAD \\
EasyHeC~\cite{EasyHeC} & $\times$ & \checkmark & $\times$ & Robot CAD and trained models \\
EasyHeC++~\cite{EasyHeCpp} & \checkmark & \checkmark & $\times$ & Robot CAD \\
Kalib~\cite{Kalib} & \checkmark & \checkmark & \checkmark & Selected robot reference point \\
\midrule
\rowcolor{black!6}
\multicolumn{5}{@{}l}{\textit{Scene-based}} \\
JCR~\cite{JCR} & \checkmark & $\times$ & $\times$ & \multirow{4}{*}{None} \\
Bi-JCR~\cite{BiJCR} & \checkmark & $\times$ & $\times$ & \\
Calib3R~\cite{Calib3R} & \checkmark & $\times$ & $\times$ & \\
\textbf{AnchorHEC (ours)} & \checkmark & \checkmark & \checkmark & \\
\bottomrule
\end{tabular}
\begin{papertablenote}
Checks indicate what the cited work demonstrates.
``Single run'' denotes calibration of both wrist and fixed cameras in one
run of a common calibration procedure, rather than one camera type at a
time. EasyHeC++ supports the two camera types separately.
\end{papertablenote}
\end{table}

We initialize the wrist transform from reconstructed relative motions,
using FK to select view pairs with diverse rotation axes because each pair
leaves translation along its rotation axis undetermined.
Because wrist-transform errors propagate to fixed-camera initialization,
we refine the wrist transform using wrist-image reprojection and metric FK constraints
through bundle adjustment (BA)~\cite{Zhi2017,Ulrich2024}.
If BA fails or retains too few inliers with the FK poses and lens
parameters held fixed, we allow FK corrections and lens refinement.
If refinement lacks sufficient image support, we return the initial wrist estimate.

We register each fixed camera to metric points triangulated from
wrist-camera poses given by the refined wrist transform and FK, reducing
reliance on reconstructed fixed-camera poses.
This requires correspondences between wrist and fixed-camera
images, which can be difficult to establish across large viewpoint differences.
We use dense matching~\cite{RoMaV2} to obtain these correspondences.
We use the number of inlier correspondences and a test for competing
camera poses to decide whether to accept registration.
If registration is not accepted, reconstruction-based estimates provide a
fallback.
\looseness=-1

We validate our approach across different reconstruction backbones in
simulation and evaluate both camera types on a physical robot.

Our contributions are:
\begin{itemize}
\interlinepenalty=10000
\item A scene-based framework for targetless calibration of wrist and
fixed cameras in a single procedure from RGB images of a common scene,
FK, and known intrinsics.

\item Kinematics-anchored wrist-camera estimation combining FK-guided pair
selection with image-based refinement to reduce the impact of
reconstruction errors.

\item Fixed-camera calibration through registration to metric scene
geometry anchored by the refined wrist transform and FK.

\end{itemize}

\section{METHOD}
\label{sec:method}

Given RGB images of a common scene from wrist and optional fixed cameras,
FK poses, and known intrinsics, AnchorHEC estimates camera-to-robot
transforms through shared reconstruction frame initialization, wrist-camera refinement, and fixed-camera
registration (Fig.~\ref{fig:overview}, Algorithm~\ref{alg:anchorhec}).
\looseness=-1

\subsection{Shared Reconstruction Frame Initialization}
\label{sec:method_start}

We call the common coordinate system of the jointly reconstructed wrist
and fixed-camera images the \emph{shared reconstruction frame}. We initialize both
camera types by placing this frame in robot coordinates.
Let $\mathbf{E}_i={}^{\mathrm{base}}\mathbf{T}_{E,i}$ map end-effector
coordinates to the robot base at wrist pose $i$. A 3DFM processes all input images to estimate
camera-to-reconstruction poses
$\hat{\mathbf{P}}_{w,i}$ for $N_w$ wrist images and
$\hat{\mathbf{P}}_{e,k,j}$ for $N_e$ images from each fixed camera $k$
($j=1,\ldots,N_e$),
with depth maps and focal lengths. The unknowns are the wrist transform
$\mathbf{X}={}^{E}\mathbf{T}_{C_w}$ (wrist camera to end effector),
the reconstruction scale $\lambda>0$, and the fixed-camera extrinsics
$\mathbf{X}_{e,k}={}^{\mathrm{base}}\mathbf{T}_{C_{e,k}}$
(fixed camera $k$ to robot base).
We assume rigid mounts, sufficient shared observations of textured
scene structure that remains stationary in the robot-base frame, and
wrist rotations about non-parallel axes~\cite{Tsai1989}.
The robot may move in view but provides no visual calibration reference.
Intrinsics, including lens distortion, are known, e.g., from factory calibration.
\looseness=-1

\paragraph{Pose relation}
Define $\mathcal{S}_{\lambda}([\mathbf{R},\mathbf{t}])=
[\mathbf{R},\lambda\mathbf{t}]$. Ideally,
$\mathbf{E}_i\mathbf{X}=\mathbf{V}\mathcal{S}_{\lambda}
(\hat{\mathbf{P}}_{w,i})$, with one reconstruction-to-base transform
$\mathbf{V}$. We first estimate $\mathbf{X}$ and $\lambda$ from relative
motions, which do not involve $\mathbf{V}$, and then fit $\mathbf{V}$
to place the fixed cameras.

\paragraph{Initial wrist transform and scale}
For a wrist-view pair $(i,j)$, let
$\mathbf{A}_{ij}=\mathbf{E}_i^{-1}\mathbf{E}_j$ and
$\mathbf{B}_{ij}=(\hat{\mathbf{P}}_{w,i})^{-1}\hat{\mathbf{P}}_{w,j}$.
Eliminating $\mathbf{V}$ gives
$\mathbf{A}_{ij}\mathbf{X}=\mathbf{X}\mathcal{S}_{\lambda}(\mathbf{B}_{ij})$,
whose translation part is
\begin{equation}
 (\mathbf{R}_{A,ij}-\mathbf{I})\mathbf{t}_X
 =\mathbf{R}_X\lambda\mathbf{t}_{B,ij}-\mathbf{t}_{A,ij}.
 \label{eq:trans_system}
\end{equation}
Each pair leaves translation along its unit rotation axis $\mathbf{u}_{ij}$
undetermined, since $(\mathbf{R}_{A,ij}-\mathbf{I})\mathbf{u}_{ij}=\mathbf{0}$.
A greedy E-optimal rule based only on FK~\cite{Kiefer1959} selects pairs
$\mathcal{I}$ by maximizing the smallest eigenvalue of
$\sum c_{ij}\mathbf{u}_{ij}\mathbf{u}_{ij}^{\top}$, an axis-coverage
heuristic rather than the translation information matrix.
Initially $c_{ij}=1$. If the smallest eigenvalue reaches 0.5, we repeat
selection with $c_{ij}=\exp(-\|\mathbf{t}_{E,i}-\mathbf{t}_{E,j}\|^2/s^2)$,
where $s$ is half the median pairwise FK distance. This penalty favors
nearby end-effector positions as a proxy for view overlap.
The number of selected pairs ranges from $N_w-1$ to $2(N_w-1)$
and depends on how unevenly the FK rotation axes are distributed.
For rank-deficient motion, we use a
rotation-weighted spanning-tree fallback. We initialize rotation with Park--Martin~\cite{Park1994}
and translation and scale with the formulation used in JCR~\cite{JCR}.
We weight both solves using $w_{ij}$, the mean of the two
frame-level confidence scores in each pair. L-BFGS~\cite{LiuNocedal1989} then minimizes
\begin{equation}
 \mathcal{L}_{\mathrm{cal}}(\mathbf{X},\lambda)=
 \frac{\displaystyle\sum_{(i,j)\in\mathcal{I}}w_{ij}
 \phi\!\left(\mathbf{A}_{ij}\mathbf{X},
       \mathbf{X}\mathcal{S}_{\lambda}(\mathbf{B}_{ij})\right)}
 {\displaystyle\sum_{(i,j)\in\mathcal{I}}w_{ij}}.
 \label{eq:cal_objective}
\end{equation}
The pose loss $\phi$ applies the Huber penalty~\cite{Huber1964}
to the norm of the normalized residual
$(\log(\mathbf{R}_1^{\top}\mathbf{R}_2)^{\vee}/\sigma_R,
(\mathbf{t}_1-\mathbf{t}_2)/\sigma_t)$,
with $\sigma_R=0.03$ rad, $\sigma_t=0.02$ m, and Huber threshold 1. The result is
$(\mathbf{X}_0,\lambda_0)$.

\paragraph{Placement in robot coordinates}
For an estimate $(\mathbf{X},\lambda)$, let
$\mathbf{C}_i=\mathbf{E}_i\mathbf{X}=[\mathbf{R}_{C,i},\mathbf{t}_{C,i}]$
and $\hat{\mathbf{P}}_{w,i}=[\mathbf{R}_{P,i},\mathbf{t}_{P,i}]$. We fit
$\mathbf{V}=[\mathbf{R}_V,\mathbf{t}_V]$ by
\begin{equation}
 \begin{aligned}
 \mathbf{R}_V&=\operatorname{Proj}_{\mathrm{SO}(3)}
   \left(\frac{1}{N_w}\sum_i\mathbf{R}_{C,i}\mathbf{R}_{P,i}^{-1}\right),\\
 \mathbf{t}_V&=\frac{1}{N_w}\sum_i
   (\mathbf{t}_{C,i}-\mathbf{R}_V\lambda\mathbf{t}_{P,i}).
 \end{aligned}
 \label{eq:vfm_base_init}
\end{equation}
This aligns the scaled reconstructed wrist trajectory with the wrist-camera
trajectory computed from FK and $\mathbf{X}$. It also places
the fixed cameras in the base frame:
\begin{equation}
 \mathbf{X}_{e,k}=\operatorname{RobustAvg}_{i}
 \left[\mathbf{V}(\mathbf{X},\lambda)
 \mathcal{S}_{\lambda}(\widetilde{\mathbf{P}}_{e,k,i})\right].
 \label{eq:external_init}
\end{equation}
Here $\widetilde{\mathbf{P}}_{e,k,i}$ is the reconstructed pose of the
fixed-camera frame assigned to wrist view $i$. The assignment uses
relative frame indices in the two image sequences, reusing fixed-camera
frames when $N_w>N_e$.
$\operatorname{RobustAvg}$ averages poses after interquartile-range
rejection of outlying positions. Fixed-camera frames need not be synchronized
with wrist images if sufficient unchanged scene structure is shared
across views. Each wrist image requires its corresponding FK pose.

\begin{algorithm}[t]
\caption{AnchorHEC}
\label{alg:anchorhec}
\scriptsize\linespread{0.95}\selectfont
\begin{algorithmic}[1]
\Require wrist images with FK poses $\mathbf{E}_i$, optional fixed-camera
frames, known intrinsics
\Ensure wrist transform $\mathbf{X}$, fixed-camera extrinsics $\mathbf{X}_{e,k}$
\Statex \textit{Shared reconstruction frame initialization} (Section~\ref{sec:method_start})
\State Reconstruct all input images with a 3DFM
\State $(\mathbf{X}_0,\lambda_0)\gets$ initial estimate from reconstructed poses and FK
  \Comment{Eqs.~\eqref{eq:trans_system}--\eqref{eq:cal_objective}}
\State $\mathbf{X}\gets\mathbf{X}_0$ and $\mathbf{X}_{e,k}\gets$ initial fixed-camera
  poses from $(\mathbf{X}_0,\lambda_0)$ \Comment{Eqs.~\eqref{eq:vfm_base_init}--\eqref{eq:external_init}}
\Statex \textit{Kinematics-anchored wrist-camera refinement} (Section~\ref{sec:method_ba})
\State Build geometrically verified SIFT tracks from wrist images only
\State \textbf{if} no tracks or fewer than 200 observations \textbf{then return}
  $\mathbf{X}$, $\mathbf{X}_{e,k}$
\State $(\mathbf{X}^{\star},r_{\mathrm{in}})\gets$ exact-FK BA from $\mathbf{X}_0$
  \Comment{Eq.~\eqref{eq:ba_objective}, $\mathbf{D}_i=\mathbf{I}$}
\State \textbf{if} exact-FK BA fails or $r_{\mathrm{in}}<0.30$ \textbf{then}
  $(\mathbf{X}^{\star},r_{\mathrm{in}})\gets$ BA with FK priors and lens refinement from $\mathbf{X}_0$
  \Comment{priors on $\mathbf{D}_i$, $\boldsymbol{\kappa}$}
\State \textbf{if} BA fails or fewer than 50 points or 200 observations survive
  \textbf{then return}
  $\mathbf{X}$, $\mathbf{X}_{e,k}$ \Comment{initial estimate}
\State $\mathbf{X}\gets\mathbf{X}^{\star}$
\Statex \textit{Fixed-camera registration} (Section~\ref{sec:method_fixed})
\For{each fixed camera $k$}
  \State $\bar{\mathbf{X}}_{e,k}\gets$ corrected shared reconstruction frame pose from $\mathbf{X}$
    \Comment{Eqs.~\eqref{eq:vfm_base_init}--\eqref{eq:external_init}, \eqref{eq:scale_refit}--\eqref{eq:range_correction}}
  \State Match fixed-camera frames to wrist views densely and triangulate metric points
    with $\mathbf{E}_i\mathbf{X}$ and factory intrinsics
  \State Fit $\mathbf{T}_1$ by PnP, then $\mathbf{T}_2$ on its outliers if supported
  \State $\mathbf{X}_{e,k}\gets\mathbf{T}_1$ \textbf{if} it has at least 30
    inliers and no ambiguity with $\mathbf{T}_2$, \textbf{else} $\bar{\mathbf{X}}_{e,k}$
\EndFor
\State \Return $\mathbf{X}$, $\mathbf{X}_{e,k}$
\end{algorithmic}
\end{algorithm}

\subsection{Kinematics-Anchored Wrist-Camera Refinement}
\label{sec:method_ba}

BA refines $\mathbf{X}_0$ using wrist-image reprojection and metric
FK constraints. Fixed-camera images influence wrist-camera estimation only through initialization.

\begin{figure}[!t]
\setlength{\abovecaptionskip}{0pt}
\centering
\setkeys{Gin}{trim={0 0 0 0},clip}
\figslot{0.8\columnwidth}{35.2mm}{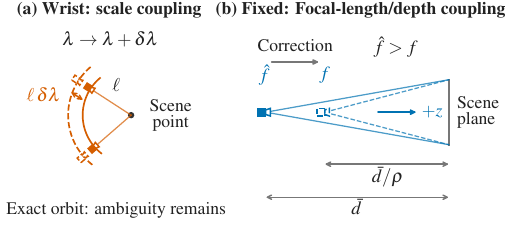}{%
  FIG. 2: CAMERA ERROR MECHANISMS\\[6pt]
  (a) Wrist: scale / axial translation\\
  Exact orbit: ambiguity remains\\[6pt]
  (b) Fixed: focal length / depth\\
  Correction direction and optical axis}
\caption{Geometric couplings. (a) Solid and dashed curves and camera
poses show solutions at $\lambda$ and $\lambda+\delta\lambda$
($\delta\lambda>0$), respectively. Scale and axial wrist translation
remain ambiguous on an exact object-centered orbit.
(b) Focal-length/depth coupling for $\hat f>f$ motivates the approximate
axial correction used in the shared reconstruction frame fallback
[Eq.~\eqref{eq:range_correction}].}
\label{fig:mechanisms}
\end{figure}

\paragraph{Scale and optical-axis coupling}
Equation~\eqref{eq:trans_system} couples scale and wrist
translation~\cite{Andreff2001}. For a camera viewing the same point at
constant distance $\ell$ in reconstruction coordinates along its optical axis $\mathbf{z}$,
$\mathbf{t}_{B,ij}=\ell(\mathbf{I}-\mathbf{R}_{B,ij})\mathbf{z}$.
Using $\mathbf{R}_{A,ij}\mathbf{R}_X=\mathbf{R}_X\mathbf{R}_{B,ij}$ gives
$(\mathbf{R}_{A,ij}-\mathbf{I})(\mathbf{t}_X+\lambda\ell\mathbf{R}_X\mathbf{z})
=-\mathbf{t}_{A,ij}$. Only the sum is determined, so scale error
$\delta\lambda$ can appear as axial translation error
$\ell\,\delta\lambda$ (Fig.~\ref{fig:mechanisms}a). Departures from this
ideal orbit provide image constraints that can help resolve the ambiguity.
An exact orbit remains ambiguous even without an explicit scale variable.

\paragraph{Camera model and objective}
Following Zhi and Schwertfeger~\cite{Zhi2017}, we refine $\mathbf{X}$
from wrist-image tracks. SIFT keypoints~\cite{Lowe2004} are matched
between all wrist-image pairs, geometrically verified, and linked into tracks.
Track $q$ has observation $\mathbf{u}_{iq}$ in view $i$. We model
$\mathbf{C}_i=\mathbf{E}_i\mathbf{D}_i\mathbf{X}$, with per-view FK
correction $\mathbf{D}_i\in\mathrm{SE}(3)$. Exact FK sets
$\mathbf{D}_i=\mathbf{I}$, using the uncorrected FK poses $\mathbf{E}_i$.
FK supplies metric translations, so no scale
variable is needed. For projection $\pi$, lens parameters
$\boldsymbol{\kappa}=(a,k_1,k_2)$ (focal multiplier $1+a$ and two
radial coefficients, with factory principal point fixed), and
base-frame points $\mathbf{p}_q$, BA minimizes
\begin{equation}
 \begin{aligned}
 \mathcal{L}_{\mathrm{BA}}=&\sum_{(i,q)\in\mathcal{O}}
 \operatorname{Huber}_{\delta}\!\left(\left\|\pi\!\left(\boldsymbol{\kappa},
 \mathbf{C}_i^{-1}\mathbf{p}_q\right)-\mathbf{u}_{iq}\right\|\right)\\
 &+\sum_i\left\|\left(\log(\mathbf{R}_{D,i})^{\vee},
 \mathbf{t}_{D,i}\right)\right\|^2_{\boldsymbol{\Sigma}_D}
 +\left\|\boldsymbol{\kappa}-\boldsymbol{\kappa}_0\right\|^2_{\boldsymbol{\Sigma}_\kappa},
 \end{aligned}
 \label{eq:ba_objective}
\end{equation}
where $\mathcal{O}$ indexes observations,
$[\mathbf{R}_{D,i},\mathbf{t}_{D,i}]=\mathbf{D}_i$, and
$\boldsymbol{\kappa}_0=(0,0,0)$ gives factory focal lengths and no
distortion. BA does not use the factory distortion
coefficients. Exact-FK BA uses 10\,px Huber and 2\,px rejection
thresholds. BA with FK priors and lens refinement uses 4\,px for both.
The prior standard deviations
are 0.3\,deg and 3\,mm for FK, and $(0.05,0.2,0.5)$ for
$\boldsymbol{\kappa}$. Each round minimizes the Huber objective with
Levenberg--Marquardt before rejecting observations and re-triangulating,
for at most ten rounds.

\paragraph{Relaxing FK and lens constraints}
BA first enforces exact FK with $\boldsymbol{\kappa}=\boldsymbol{\kappa}_0$.
The retained observation ratio $r_{\mathrm{in}}$ is the fraction of
input-track observations that survive. If exact-FK BA
fails or $r_{\mathrm{in}}<0.30$, it restarts from $\mathbf{X}_0$ with
Gaussian priors on the FK corrections~\cite{Koide2019,Ulrich2024} and weak priors
on $\boldsymbol{\kappa}$.
This relaxation allows BA to adjust FK poses and lens parameters 
to fit the image tracks and raises the rejection threshold from 2 to 4\,px.
Freeing these variables can reduce
accuracy when FK poses and lens parameters are reliable. We therefore start
with exact FK (Section~\ref{sec:exp_components}(b)). A minimum-support
check requires the selected solution to retain at least 50 points and 200 observations. If it
fails this check, or the input tracks contain fewer than 200
observations, we return the initial estimates for both camera types
and skip registration. Wrist-transform errors propagate to the geometry
used for registration, and the registration acceptance rule does not independently
assess wrist-transform accuracy.

\subsection{Fixed-Camera Registration to Wrist-Anchored Geometry}
\label{sec:method_fixed}

\paragraph{Metric registration}
To estimate the fixed-camera extrinsics, each fixed camera is registered to
triangulated scene points, as in structure-based localization~\cite{HLoc}.
During registration, we hold the refined wrist transform $\mathbf{X}^{\star}$
fixed and triangulate metric points using factory intrinsics and the
wrist-camera poses $\mathbf{E}_i\mathbf{X}^{\star}$, with uncorrected FK poses $\mathbf{E}_i$.

Dense matching~\cite{RoMaV2} links each fixed-camera
image to each wrist image. Unlike BA, registration undistorts correspondences with the factory
coefficients and verifies them geometrically per image pair. For each fixed-camera image pixel matched in at least two wrist views,
we triangulate a metric base-frame point. We retain it if its depth is
positive in all observing wrist cameras, the maximum angle between
the lines containing the viewing rays is at least 2\,deg, and its mean reprojection error
in the undistorted wrist images is at most 6\,px.
We pool the 2D--3D correspondences across the $N_e$ frames of each fixed
camera. A perspective-$n$-point (PnP) solver with RANSAC~\cite{Fischler1981} and nonlinear refinement
estimates a first pose $\mathbf{T}_1$ with a 3\,px inlier threshold.

\paragraph{Ambiguity test and acceptance rule}
A large PnP inlier set can support a wrong pose on repetitive structure. We fit a second PnP pose $\mathbf{T}_2$ to the
correspondences rejected by $\mathbf{T}_1$. Each fit requires at least 12 inliers. We reject
the registration if a second pose is found with at least one fifth as
many inliers as the first pose and the two camera centers are at least
0.5\,m apart. Otherwise, the first pose requires at least 30 inliers for acceptance. Rejected or unsupported registrations fall back to the
corrected shared reconstruction frame pose $\bar{\mathbf{X}}_{e,k}$, defined below. The second pose is
never returned.

\paragraph{Shared reconstruction frame fallback}
BA estimates metric $\mathbf{X}^{\star}$ but no reconstruction scale.
For Eq.~\eqref{eq:external_init}, we refit the reconstruction scale
to the wrist-camera centers
$\mathbf{t}_{C,i}$ of $\mathbf{E}_i\mathbf{X}^{\star}$, using
$\mathbf{R}_V$ recomputed from $\mathbf{X}^{\star}$ by
Eq.~\eqref{eq:vfm_base_init}:
\begin{equation}
 \lambda^{\star}=\frac{\sum_i(\mathbf{t}_{C,i}-\bar{\mathbf{t}}_C)^{\top}
 \mathbf{R}_V(\mathbf{t}_{P,i}-\bar{\mathbf{t}}_P)}
 {\sum_i\|\mathbf{t}_{P,i}-\bar{\mathbf{t}}_P\|^2}.
 \label{eq:scale_refit}
\end{equation}
Here $\bar{\mathbf{t}}_C$ and $\bar{\mathbf{t}}_P$ are means over
wrist views. Assuming focal-length error induces axial
displacement~\cite{HartleyZisserman2003}, we apply an approximate
axial correction to the translation $\mathbf{t}_{e,k}$ of $\mathbf{X}_{e,k}$
along the unit optical axis $\mathbf{z}_k$ expressed in
the robot base frame:
\begin{equation}
 \mathbf{t}_{e,k}\leftarrow\mathbf{t}_{e,k}+
 \lambda^{\star}\bar d_k\left(1-\rho_k^{-1}\right)\mathbf{z}_k.
 \label{eq:range_correction}
\end{equation}
The focal ratio $\rho_k$ is the median over frames of the
predicted-to-factory focal-length ratio, averaged over both image
axes after rescaling predictions to the input resolution. Thus,
$\rho_k=\hat f/f$ for isotropic focal error (Fig.~\ref{fig:mechanisms}b).
For each frame, we take the median depth of high-confidence points in
reconstruction coordinates. We then take the median of these frame-level
values to obtain $\bar d_k$. The resulting corrected shared reconstruction frame pose
$\bar{\mathbf{X}}_{e,k}$ needs no wrist-to-fixed correspondences.
The correction's effect depends on reconstruction errors
(Section~\ref{sec:exp_components}(c)).
AnchorHEC also reports the path of each estimate (exact-FK BA, BA with
FK priors and lens refinement, or the initial estimate for the wrist camera,
and an accepted registration or a fallback for each fixed camera), so
fallbacks can be identified without ground truth.

\section{EXPERIMENTS}
\label{sec:experiments}
\label{sec:exp_setup}

We evaluate calibration accuracy for both camera types and sensitivity
to the initial reconstruction. Main comparisons use Algorithm~\ref{alg:anchorhec}'s
parameters and decision rules without dataset-specific adjustments.
\looseness=-1

\begin{figure}[!t]
\centering
\includegraphics[width=\columnwidth]{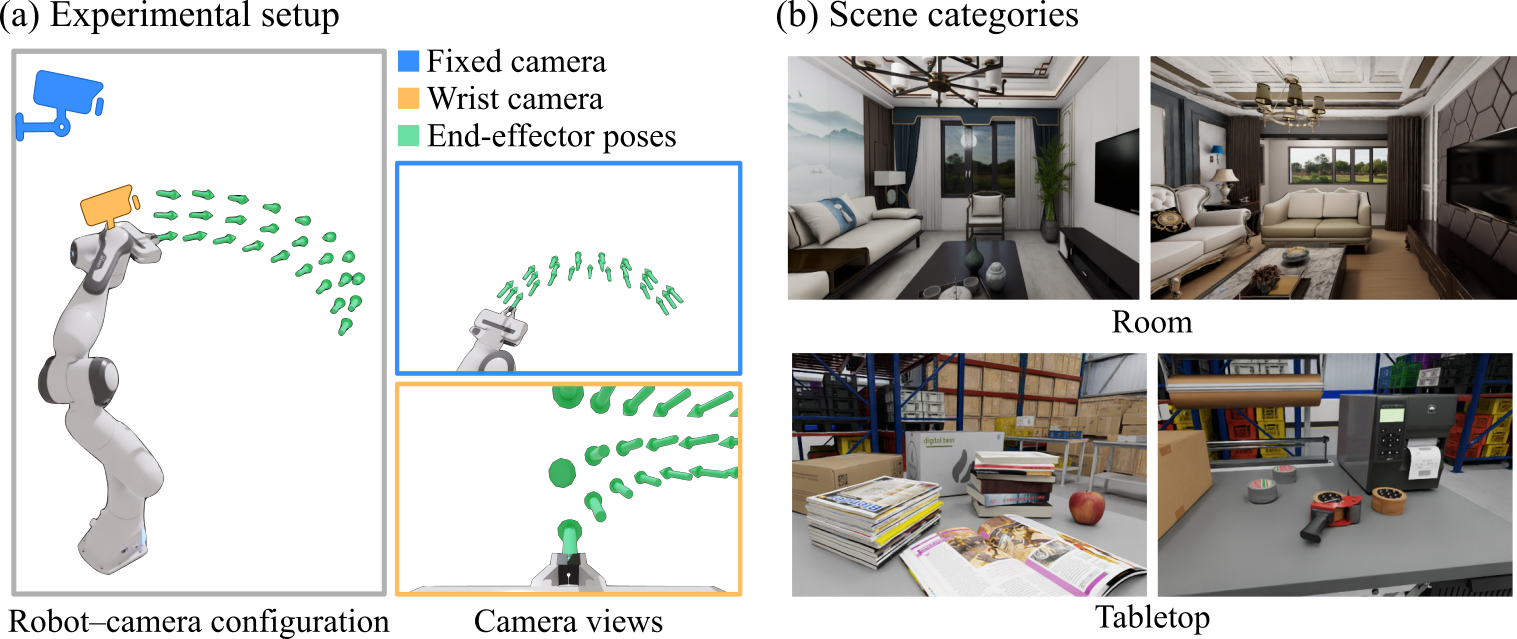}
\caption{Simulation setup. (a) Robot--camera configuration, the
multi-height trajectory (sampled end-effector poses in green), and views
from the fixed (blue) and wrist (orange) cameras. (b) Room and tabletop
scenes.}
\label{fig:simulation}
\end{figure}

\paragraph*{Evaluation protocol}
Unless stated otherwise, we used five matched sampling trials with
$N_w=15$ wrist poses and $N_e=5$ frames from the same viewpoint per
fixed camera. When all available wrist poses were selected, each trial used
the same wrist-image set.
The 3DFM backbones are DUSt3R, MASt3R, and VGGT,
with VGGT as the default outside Table~\ref{tab:main}.
ViPE~\cite{VIPE}, a video reconstruction system, serves as
an alternative source of the initial reconstruction.
All methods used matched wrist-image subsets, without robot masking.
For each backbone, JCR and AnchorHEC also used the same fixed-camera
images and reconstruction output. 
With $N_w=15$ and $N_e=5$, BA took about one minute at $640{\times}360$
and two minutes at $1920{\times}1080$ per subset on one CPU core.
At these resolutions, dense matching with the RoMa~v2 base setting~\cite{RoMaV2}
took about 12 and 19\,s per fixed camera, respectively, on one GPU,
excluding triangulation and PnP.
\looseness=-1

\paragraph*{Baselines}
JCR~\cite{JCR} uses its original adjacent-frame, scale-aware
$\mathbf{A}\mathbf{X}=\mathbf{X}\mathbf{B}$ solver with reconstructed poses and intrinsics. Table~\ref{tab:exp_ext_wrist}
also reports its accuracy on wrist-only reconstructions ($M=0$).
To obtain the fixed-camera extrinsics absent from prior scene-based
methods (Table~\ref{tab:related_compare}), we apply shared reconstruction
frame placement (SF), Eqs.~\eqref{eq:vfm_base_init}--\eqref{eq:external_init}, to
JCR's wrist transform and scale, giving \emph{JCR~+~SF}.
Table~\ref{tab:wrist_ablation} also applies our registration to the JCR and
initial wrist estimates. Our \emph{initial estimate} differs from JCR through FK-selected pairs,
confidence-based pair weights, and Eq.~\eqref{eq:cal_objective}'s robust
objective. Both use reconstructed poses and FK without axial correction
or registration. The baseline of Zhi and Schwertfeger~\cite{Zhi2017}, whose BA formulation we
follow, uses
the authors' code and exhaustive matcher. Known intrinsics and its own
reconstruction make it independent of 3DFMs. It estimates only the wrist
transform, without fallback.
We compare scene-based methods without robot-specific visual references.
Robot-body-based methods require visible robot features plus CAD models,
trained models, or a selected reference point
(Table~\ref{tab:related_compare}).
Calib3R~\cite{Calib3R} has no public implementation.

\paragraph*{Metrics and statistics}
Translation error $e_t$ and geodesic rotation error $e_\theta$
measure accuracy against ground truth (GT) or the physical reference,
with one run defined by a wrist subset (one sample of $N_w$ wrist poses) and a fixed-camera
subset. Unless stated otherwise, fixed-camera errors are
averaged within each run and then over runs. Brackets are 95\% paired-bootstrap intervals for the stated
differences or relative reductions
(20,000 draws, stratified by scene and wrist-pose budget). We resampled
wrist subsets, not scenes, keeping all runs of a wrist subset
together. An absolute reduction is the comparison method's error minus ours.
A relative reduction is this difference divided by that method's error.
We average relative reductions across the stated conditions.
\looseness=-1

\subsection{Mixed-Camera Calibration in Simulation}
\label{sec:exp_simulation}
\label{sec:exp_main}

We vary the reconstruction backbone, wrist-pose budget, and fixed-camera
subset.

\begin{table*}[!t]
\caption{Simulation Accuracy Across Reconstruction Backbones}
\label{tab:main}
\centering
\scriptsize
\setlength{\tabcolsep}{2.6pt}
\renewcommand{\arraystretch}{0.95}
\begin{tabular}{@{}lc
  >{\columncolor{black!6}}r r >{\columncolor{black!6}}r r >{\columncolor{black!6}}r r !{\hspace{3pt}} >{\columncolor{black!6}}r r >{\columncolor{black!6}}r r >{\columncolor{black!6}}r r
  !{\hspace{8pt}}
  >{\columncolor{black!6}}r r >{\columncolor{black!6}}r r !{\hspace{3pt}} >{\columncolor{black!6}}r r >{\columncolor{black!6}}r r@{}}
\toprule
& & \multicolumn{12}{c}{Wrist camera} & \multicolumn{8}{c}{Fixed camera} \\
\cmidrule(lr){3-14}\cmidrule(l){15-22}
& & \multicolumn{6}{c}{Room} & \multicolumn{6}{c}{Tabletop}
  & \multicolumn{4}{c}{Room} & \multicolumn{4}{c}{Tabletop} \\
\cmidrule(lr){3-8}\cmidrule(lr){9-14}\cmidrule(lr){15-18}\cmidrule(l){19-22}
Backbone & $N_w$
& \multicolumn{2}{c}{JCR~\cite{JCR}} & \multicolumn{2}{c}{Zhi~\cite{Zhi2017}} & \multicolumn{2}{c}{\textbf{Ours}}
& \multicolumn{2}{c}{JCR~\cite{JCR}} & \multicolumn{2}{c}{Zhi~\cite{Zhi2017}} & \multicolumn{2}{c}{\textbf{Ours}}
& \multicolumn{2}{c}{JCR~\cite{JCR} + SF} & \multicolumn{2}{c}{\textbf{Ours}}
& \multicolumn{2}{c}{JCR~\cite{JCR} + SF} & \multicolumn{2}{c}{\textbf{Ours}} \\
\cmidrule(lr){3-4}\cmidrule(lr){5-6}\cmidrule(lr){7-8}\cmidrule(lr){9-10}\cmidrule(lr){11-12}\cmidrule(lr){13-14}
\cmidrule(lr){15-16}\cmidrule(lr){17-18}\cmidrule(lr){19-20}\cmidrule(l){21-22}
& & $e_t$ & $e_\theta$ & $e_t$ & $e_\theta$ & $e_t$ & $e_\theta$ & $e_t$ & $e_\theta$ & $e_t$ & $e_\theta$ & $e_t$ & $e_\theta$
  & $e_t$ & $e_\theta$ & $e_t$ & $e_\theta$ & $e_t$ & $e_\theta$ & $e_t$ & $e_\theta$ \\
\midrule
DUSt3R~\cite{DUSt3R} & 5 & 40.79 & 3.31 & 1154.79$^\dagger$ & 6.78 & \textbf{7.72} & \textbf{0.96} & 29.30 & 3.35 & \multicolumn{2}{c}{Fail (0/10)} & \textbf{0.69} & \textbf{0.31} & 50.20 & 2.87 & \textbf{8.35} & \textbf{1.09} & 43.62 & 3.39 & \textbf{1.27} & \textbf{0.47} \\
 & 15 & 66.43 & 2.67 & 64.37$^\dagger$ & 4.48 & \textbf{1.01} & \textbf{0.22} & 40.30 & 2.86 & 1.75 & 0.65 & \textbf{0.18} & \textbf{0.10} & 95.98 & 2.46 & \textbf{1.78} & \textbf{0.32} & 65.25 & 2.98 & \textbf{0.61} & \textbf{0.19} \\
 & 25 & 83.96 & 2.06 & 6.56 & 0.76 & \textbf{0.79} & \textbf{0.21} & 61.28 & 1.93 & 0.84 & 0.70 & \textbf{0.16} & \textbf{0.09} & 124.36 & 1.90 & \textbf{1.47} & \textbf{0.32} & 119.34 & 1.60 & \textbf{0.55} & \textbf{0.18} \\
\midrule
MASt3R~\cite{MASt3R} & 5 & 11.57 & 1.83 & 1154.79$^\dagger$ & 6.78 & \textbf{4.44} & \textbf{0.81} & 4.50 & 2.06 & \multicolumn{2}{c}{Fail (0/10)} & \textbf{0.69} & \textbf{0.31} & 14.77 & 2.37 & \textbf{4.87} & \textbf{0.89} & 11.64 & 2.23 & \textbf{1.27} & \textbf{0.47} \\
 & 15 & 8.61 & 0.93 & 64.37$^\dagger$ & 4.48 & \textbf{1.02} & \textbf{0.22} & 6.42 & 1.86 & 1.75 & 0.65 & \textbf{0.19} & \textbf{0.10} & 7.67 & 1.02 & \textbf{1.71} & \textbf{0.31} & 6.03 & 1.84 & \textbf{0.65} & \textbf{0.19} \\
 & 25 & 9.08 & 0.83 & 6.56 & 0.76 & \textbf{0.82} & \textbf{0.18} & 6.94 & 1.73 & 0.84 & 0.70 & \textbf{0.16} & \textbf{0.09} & 9.71 & 0.83 & \textbf{1.38} & \textbf{0.29} & 7.05 & 1.63 & \textbf{0.55} & \textbf{0.18} \\
\midrule
VGGT~\cite{VGGT} & 5 & 6.32 & 0.47 & 1154.79$^\dagger$ & 6.78 & \textbf{3.02} & \textbf{0.31} & 3.33 & 0.67 & \multicolumn{2}{c}{Fail (0/10)} & \textbf{0.65} & \textbf{0.30} & 9.33 & 0.64 & \textbf{3.84} & \textbf{0.51} & 7.44 & 0.61 & \textbf{1.18} & \textbf{0.45} \\
 & 15 & 6.40 & 0.34 & 64.37$^\dagger$ & 4.48 & \textbf{1.02} & \textbf{0.22} & 2.98 & 0.41 & 1.75 & 0.65 & \textbf{0.19} & \textbf{0.10} & 4.26 & 0.56 & \textbf{1.79} & \textbf{0.32} & 4.47 & 0.57 & \textbf{0.66} & \textbf{0.19} \\
 & 25 & 6.33 & 0.40 & 6.56 & 0.76 & \textbf{0.79} & \textbf{0.20} & 2.66 & 0.46 & 0.84 & 0.70 & \textbf{0.16} & \textbf{0.09} & 4.78 & 0.62 & \textbf{1.40} & \textbf{0.29} & 4.56 & 0.56 & \textbf{0.55} & \textbf{0.18} \\
\midrule
ViPE~\cite{VIPE} & 5 & 5.33 & 0.65 & 1154.79$^\dagger$ & 6.78 & \textbf{2.49} & \textbf{0.46} & 4.56 & 1.14 & \multicolumn{2}{c}{Fail (0/10)} & \textbf{0.69} & \textbf{0.31} & 117.17 & 1.63 & \textbf{55.52} & \textbf{1.00} & 65.71 & 3.60 & \textbf{1.26} & \textbf{0.49} \\
 & 15 & 2.05 & 0.29 & 64.37$^\dagger$ & 4.48 & \textbf{1.02} & \textbf{0.22} & 3.31 & 0.63 & 1.75 & 0.65 & \textbf{0.18} & \textbf{0.10} & 114.55 & 1.41 & \textbf{1.75} & \textbf{0.32} & 66.25 & 5.30 & \textbf{0.67} & \textbf{0.20} \\
 & 25 & 2.14 & 0.37 & 6.56 & 0.76 & \textbf{0.82} & \textbf{0.19} & 5.35 & 0.36 & 0.84 & 0.70 & \textbf{0.16} & \textbf{0.09} & 115.22 & 1.16 & \textbf{1.38} & \textbf{0.29} & 58.55 & 5.74 & \textbf{0.55} & \textbf{0.18} \\
\bottomrule
\end{tabular}
\begin{papertablenote}
Mean $e_t$ (cm)/$e_\theta$ (deg) over wrist subsets with an output, lowest in bold.
The baseline of Zhi and Schwertfeger does not use a 3DFM, so its entries repeat across backbones.
Fail (0/10): no output on any of the ten wrist subsets. $^\dagger$This baseline returns outputs on 2/10 and 9/10 room wrist subsets at $N_w=5$ and 15,
with maximum translation errors of 2300.02 and 421.82\,cm, respectively.
At $N_w=15$, its median errors are 5.49\,cm/1.19\,deg.
$N_w$: number of sampled wrist poses (one image per pose).
Two scenes per type, five trials per scene and $N_w$, and one top
fixed camera ($M=1$, $N_e=5$ images).
The large ViPE room fixed-camera error at $N_w=5$ reflects inaccurate initial
estimates retained on five of ten wrist subsets, for which registration was skipped.
\end{papertablenote}
\end{table*}

\begin{figure}[!t]
\centering
\figslot{0.8\linewidth}{52mm}{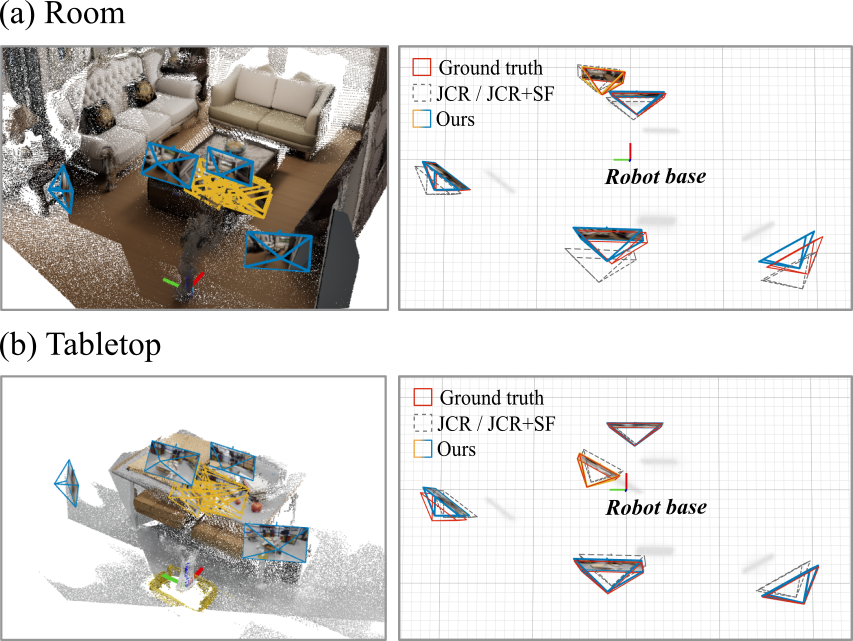}{%
  SIMULATION RESULTS\par
  USER VIEWER CAPTURE\par\medskip
  Room result / Tabletop result\par
  Top: pointmap and calibrated cameras\par
  Bottom: GT / JCR + SF / AnchorHEC\par\medskip
  $N_w=15$, $M=4$\par
  sim-room (seed 7220)\par
  sim-tabletop (seed 5334)}
\captionof{figure}{Qualitative simulation results for (a) room and
(b) tabletop runs with $N_w=15$ and $M=4$ (Table~\ref{tab:exp_ext_wrist}).
Left: the 3DFM pointmap placed in the robot base frame with our wrist
(orange) and fixed (blue) cameras. Right: GT and estimated poses for one
wrist view and four fixed cameras, comparing AnchorHEC with JCR for the
wrist camera and JCR~+~SF for the fixed cameras.}
\label{fig:simulation_results}
\end{figure}

\paragraph*{Setup}
NVIDIA Isaac Sim provides Franka Panda poses and GT extrinsics for
25-pose trajectories aimed at a scene point
(Fig.~\ref{fig:simulation}a). Each trajectory is a multi-height orbit
at three heights. The room scenes are drawn from InteriorAgent~\cite{InteriorAgent2025}.
The tabletop scenes contain textured objects within the robot's reach
(Fig.~\ref{fig:simulation}b).
Fixed-camera viewpoints are front,
top, left-side, and right-side. Left and right are defined from the
robot's perspective facing the workspace. The top camera looks obliquely
down from a higher position. All simulated views
are distortion-free $640{\times}360$ images.

\begin{table}[!t]
\begin{minipage}[t]{\columnwidth}\vspace{0pt}
\captionof{table}{Calibration Accuracy versus the Number of Fixed Cameras}
\label{tab:exp_ext_wrist}
\setlength{\aboverulesep}{1pt}
\setlength{\belowrulesep}{1.5pt}
\centering
\scriptsize
\setlength{\tabcolsep}{3.2pt}
\renewcommand{\arraystretch}{0.8}
\begin{minipage}{\linewidth}
\begin{tabular*}{\linewidth}{@{\extracolsep{\fill}}c >{\columncolor{black!6}}r r >{\columncolor{black!6}}r r !{\hspace{3pt}} >{\columncolor{black!6}}r r >{\columncolor{black!6}}r r@{}}
\toprule
& \multicolumn{8}{c}{Wrist camera} \\
\cmidrule(l){2-9}
& \multicolumn{4}{c}{Room} & \multicolumn{4}{c}{Tabletop} \\
\cmidrule(lr){2-5}\cmidrule(l){6-9}
\multirow{2}{*}{$M$} & \multicolumn{2}{c}{JCR~\cite{JCR}} & \multicolumn{2}{c}{\textbf{Ours}} & \multicolumn{2}{c}{JCR~\cite{JCR}} & \multicolumn{2}{c}{\textbf{Ours}} \\
\cmidrule(lr){2-3}\cmidrule(lr){4-5}\cmidrule(lr){6-7}\cmidrule(lr){8-9}
& $e_t$ & $e_\theta$ & $e_t$ & $e_\theta$ & $e_t$ & $e_\theta$ & $e_t$ & $e_\theta$ \\
\midrule
0 & 2.68 & 0.33 & \textbf{1.06} & \textbf{0.22} & 2.94 & 0.70 & \textbf{0.21} & \textbf{0.11} \\
1 & 4.60 & 0.34 & \textbf{1.02} & \textbf{0.22} & 3.25 & 0.50 & \textbf{0.19} & \textbf{0.10} \\
2 & 5.20 & 0.33 & \textbf{1.02} & \textbf{0.22} & 3.91 & 0.47 & \textbf{0.19} & \textbf{0.10} \\
3 & 5.64 & 0.34 & \textbf{1.02} & \textbf{0.22} & 4.38 & 0.49 & \textbf{0.19} & \textbf{0.10} \\
4 & 6.53 & 0.39 & \textbf{1.06} & \textbf{0.22} & 4.67 & 0.52 & \textbf{0.21} & \textbf{0.11} \\
\bottomrule
\end{tabular*}
\par\vspace{5pt}
\begin{tabular*}{\linewidth}{@{\extracolsep{\fill}}c >{\columncolor{black!6}}w{r}{20pt}@{\extracolsep{0pt}\hspace{4pt}}w{r}{20pt}@{\extracolsep{\fill}\hspace{6pt}} >{\columncolor{black!6}}w{r}{20pt}@{\extracolsep{0pt}\hspace{4pt}}w{r}{20pt}@{\extracolsep{\fill}\hspace{6pt}} >{\columncolor{black!6}}w{r}{20pt}@{\extracolsep{0pt}\hspace{4pt}}w{r}{20pt}@{\extracolsep{\fill}\hspace{6pt}} >{\columncolor{black!6}}w{r}{20pt}@{\extracolsep{0pt}\hspace{4pt}}w{r}{20pt}@{}}
& \multicolumn{8}{c}{Fixed camera} \\
\cmidrule(l){2-9}
& \multicolumn{4}{c}{Room} & \multicolumn{4}{c}{Tabletop} \\
\cmidrule(lr){2-5}\cmidrule(l){6-9}
\multirow{2}{*}{$M$} & \multicolumn{2}{c}{JCR~\cite{JCR} + SF} & \multicolumn{2}{c}{\textbf{Ours}} & \multicolumn{2}{c}{JCR~\cite{JCR} + SF} & \multicolumn{2}{c}{\textbf{Ours}} \\
\cmidrule(lr){2-3}\cmidrule(lr){4-5}\cmidrule(lr){6-7}\cmidrule(lr){8-9}
& $e_t$ & $e_\theta$ & $e_t$ & $e_\theta$ & $e_t$ & $e_\theta$ & $e_t$ & $e_\theta$ \\
\midrule
1 & 5.93 & 0.53 & \textbf{1.82} & \textbf{0.28} & 10.96 & 0.68 & \textbf{6.32} & \textbf{0.30} \\
2 & 4.80 & 0.49 & \textbf{1.82} & \textbf{0.28} & 8.49 & 0.60 & \textbf{3.61} & \textbf{0.26} \\
3 & 4.60 & 0.49 & \textbf{1.82} & \textbf{0.28} & 7.23 & 0.58 & \textbf{2.22} & \textbf{0.28} \\
4 & 4.42 & 0.52 & \textbf{1.78} & \textbf{0.27} & 6.06 & 0.56 & \textbf{1.43} & \textbf{0.29} \\
\bottomrule
\end{tabular*}

\begin{papertablenote}
Mean $e_t$ (cm)/$e_\theta$ (deg), lower in bold.
$M$: number of fixed cameras jointly reconstructed with the wrist images.
VGGT, $N_w=15$ wrist poses and $N_e=5$ images per fixed camera.
Two scenes per type, all size-$M$ subsets of four cameras, and five
wrist-sampling trials per subset (20 for $M=0,4$).
$M=0$: wrist-only reconstructions.
\end{papertablenote}
\end{minipage}
\end{minipage}
\end{table}

\paragraph{Calibration accuracy across backbones}

Table~\ref{tab:main} compares wrist and fixed-camera accuracy across
three 3DFMs and the additional ViPE backbone at three wrist-pose budgets.
In all 24 conditions, our mean wrist
translation and rotation errors are lower than JCR's, by 84.6\% and 70.0\% on
average. Applying our FK-guided pair selection to the same JCR solver
reduces mean wrist translation error by 33.3\% on average across all 24 conditions.
Error decreases in 23 conditions, with ViPE in tabletop scenes at $N_w=15$
as the sole exception. With DUSt3R, JCR's error grows with $N_w$: its scale estimate
falls increasingly below the true scale as denser sampling makes
adjacent-frame pairs rotate less, and DUSt3R's reconstructed translation directions between adjacent views
deviate more from GT than those of the other backbones. The baseline of Zhi and Schwertfeger returns an output on only two
of ten room wrist subsets and on no tabletop wrist subset at $N_w=5$, and its errors
exceed ours at $N_w=15$ and 25.

\begin{figure*}[!t]
\centering
\figslot{\textwidth}{31mm}{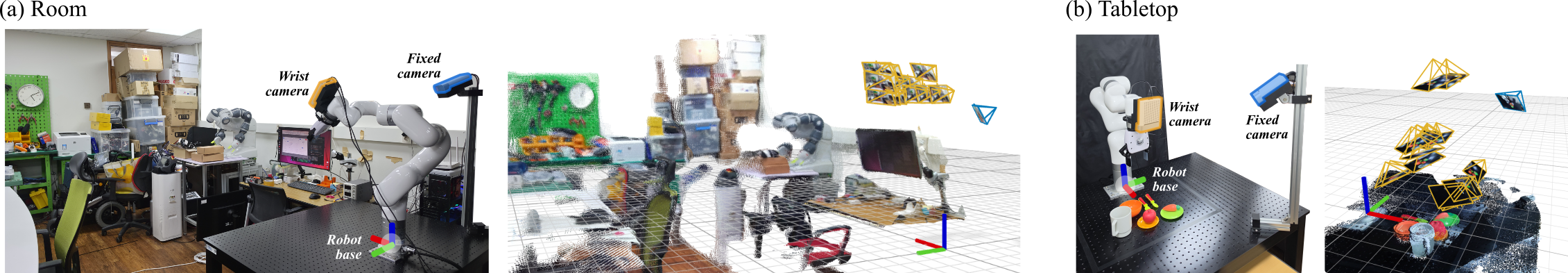}{%
  PHYSICAL-ROBOT VALIDATION, USER ARTWORK SLOT\par\medskip
  (a) Room / tabletop camera mounts\par
  (b) Wrist / fixed input views\par
  (c) Pointmap and base-frame camera poses: reference / JCR + SF / AnchorHEC\par\medskip
  Viewer cases physical-room and physical-tabletop\par
  $N_w=15$, $N_e=5$}
\caption{Physical-robot setup and qualitative results in the (a) room and
(b) tabletop acquisitions. Left: the robot with the wrist and fixed cameras.
Right: the 3DFM pointmap placed in the robot base frame with our wrist
(orange) and fixed (blue) cameras for an example $N_w=15$ subset from
Table~\ref{tab:physical}.}
\label{fig:physical}
\end{figure*}

At $N_w=15$ and $25$, the four backbones give nearly identical wrist
errors despite their different initial estimates. The exact-FK objective
is shared across backbones, which supply different starting estimates.
Across the same 24 conditions, mean wrist translation error falls by
73.3\% relative to the initial estimate, improving in 23 conditions.
The exception is ViPE at $N_w=15$ in room scenes, where BA increases
the mean error from 0.89 to 1.02\,cm.

Fixed-camera translation error is also lower than that of JCR~+~SF
in every condition. At $N_w=15$ and $25$, fixed-camera errors are
similar across backbones. Across all three wrist-pose budgets, BA succeeds
on 55 of 60 runs per backbone. All registrations on these runs are
accepted, with mean fixed-camera errors of 1.29 to 1.31\,cm.

\paragraph{Sensitivity to fixed-camera count}

In Table~\ref{tab:exp_ext_wrist}, JCR's wrist translation error grows with
$M$ in both scene types, while ours remains nearly constant. JCR uses the
wrist-camera poses from each joint reconstruction. Fixed-camera images affect only the
initialization of our BA, not its objective.
With wrist-only reconstruction ($M=0$), our mean wrist errors differ from
those at $M=1$--4 by less than 0.05\,cm and at most 0.01\,deg in each scene type.
Fixed-camera images are therefore optional for wrist calibration.

Relative to JCR~+~SF, fixed-camera translation error falls
by 3.11\,cm in room scenes and 4.80\,cm in tabletop scenes
(Fig.~\ref{fig:simulation_results}).
Fixed-camera error still varies with $M$, especially in tabletop
scenes, even though each camera appears in the same number of runs
for a given $M$. The variation is dominated by one camera whose registration is
consistently rejected: its mean fallback error decreases from 45.71\,cm
at $M=1$ to 6.23\,cm at $M=4$ (Section~\ref{sec:exp_components}(e)).

\subsection{Physical-Robot Validation}
\label{sec:exp_real}

We applied the same procedure and settings to two physical acquisitions
(Table~\ref{tab:physical}, Fig.~\ref{fig:physical}).

\begin{table}[!t]
\caption{Physical-Robot Evaluation}
\label{tab:physical}
\setlength{\aboverulesep}{1pt}
\setlength{\belowrulesep}{1.5pt}
\centering
\scriptsize
\renewcommand{\arraystretch}{0.9}
\begin{minipage}{\linewidth}
\setlength{\tabcolsep}{1.2pt}
\begin{tabular*}{\linewidth}{@{\extracolsep{\fill}}c >{\columncolor{black!6}}r r >{\columncolor{black!6}}r r >{\columncolor{black!6}}r r !{\hspace{3pt}} >{\columncolor{black!6}}r r >{\columncolor{black!6}}r r >{\columncolor{black!6}}r r@{}}
\toprule
& \multicolumn{12}{c}{Wrist camera} \\
\cmidrule(l){2-13}
& \multicolumn{6}{c}{Room} & \multicolumn{6}{c}{Tabletop} \\
\cmidrule(lr){2-7}\cmidrule(l){8-13}
\multirow{2}{*}{$N_w$} & \multicolumn{2}{c}{JCR~\cite{JCR}} & \multicolumn{2}{c}{Zhi~\cite{Zhi2017}} & \multicolumn{2}{c}{\textbf{Ours}} & \multicolumn{2}{c}{JCR~\cite{JCR}} & \multicolumn{2}{c}{Zhi~\cite{Zhi2017}} & \multicolumn{2}{c}{\textbf{Ours}} \\
\cmidrule(lr){2-3}\cmidrule(lr){4-5}\cmidrule(lr){6-7}\cmidrule(lr){8-9}\cmidrule(lr){10-11}\cmidrule(lr){12-13}
& $e_t$ & $e_\theta$ & $e_t$ & $e_\theta$ & $e_t$ & $e_\theta$ & $e_t$ & $e_\theta$ & $e_t$ & $e_\theta$ & $e_t$ & $e_\theta$ \\
\midrule
5 & 10.81 & 4.18 & 18.55 & 5.91 & \textbf{3.50} & \textbf{1.62} & 13.24 & 14.07 & \multicolumn{2}{c}{Fail (0/5)} & \textbf{6.10} & \textbf{2.37} \\
15 & 10.16 & 4.13 & 10.61 & 4.32 & \textbf{1.35} & \textbf{1.48} & 2.41 & 1.83 & 1.75 & \textbf{1.09} & \textbf{1.22} & 1.11 \\
25 & 11.18 & 4.31 & 11.64 & 4.16 & \textbf{1.33} & \textbf{1.37} & 1.90 & 1.95 & 1.92 & 1.17 & \textbf{1.32} & \textbf{1.00} \\
\bottomrule
\end{tabular*}
\par\vspace{5pt}
\begin{tabular*}{\linewidth}{@{\extracolsep{\fill}}c >{\columncolor{black!6}}w{r}{20pt}@{\extracolsep{0pt}\hspace{4pt}}w{r}{20pt}@{\extracolsep{\fill}\hspace{6pt}} >{\columncolor{black!6}}w{r}{20pt}@{\extracolsep{0pt}\hspace{4pt}}w{r}{20pt}@{\extracolsep{\fill}\hspace{6pt}} >{\columncolor{black!6}}w{r}{20pt}@{\extracolsep{0pt}\hspace{4pt}}w{r}{20pt}@{\extracolsep{\fill}\hspace{6pt}} >{\columncolor{black!6}}w{r}{20pt}@{\extracolsep{0pt}\hspace{4pt}}w{r}{20pt}@{}}
& \multicolumn{8}{c}{Fixed camera} \\
\cmidrule(l){2-9}
& \multicolumn{4}{c}{Room} & \multicolumn{4}{c}{Tabletop} \\
\cmidrule(lr){2-5}\cmidrule(l){6-9}
\multirow{2}{*}{$N_w$} & \multicolumn{2}{c}{JCR~\cite{JCR} + SF} & \multicolumn{2}{c}{\textbf{Ours}} & \multicolumn{2}{c}{JCR~\cite{JCR} + SF} & \multicolumn{2}{c}{\textbf{Ours}} \\
\cmidrule(lr){2-3}\cmidrule(lr){4-5}\cmidrule(lr){6-7}\cmidrule(lr){8-9}
& $e_t$ & $e_\theta$ & $e_t$ & $e_\theta$ & $e_t$ & $e_\theta$ & $e_t$ & $e_\theta$ \\
\midrule
5 & 13.41 & 3.03 & \textbf{5.78} & \textbf{2.45} & 16.95 & 14.66 & \textbf{7.95} & \textbf{4.50} \\
15 & 8.76 & 3.22 & \textbf{1.63} & \textbf{1.40} & 3.42 & 2.40 & \textbf{1.75} & \textbf{1.46} \\
25 & 8.19 & 3.00 & \textbf{1.66} & \textbf{1.43} & 3.18 & 2.20 & \textbf{1.88} & \textbf{1.24} \\
\bottomrule
\end{tabular*}
\begin{papertablenote}
$e_t$ (cm)/$e_\theta$ (deg), lowest per scene and budget in bold.
$N_w$: number of sampled wrist poses (one image per pose).
VGGT for JCR and Ours, one fixed camera ($M=1$, $N_e=5$ images),
means over five wrist subsets.
Fail (0/5): no output on any of the five subsets.
At $N_w=25$, room subsets reuse all 25 poses, while tabletop subsets
sample 25 poses at random. At $N_w=5$, four of five tabletop subsets
returned the initial estimates for both camera types without attempting
registration.
\end{papertablenote}
\end{minipage}
\end{table}

\paragraph*{Setup}
A physical xArm6 carried a wrist-mounted Azure Kinect, and a fixed
Azure Kinect observed the scene (Fig.~\ref{fig:physical}). Both provided $1920{\times}1080$ images
and factory intrinsics with eight distortion coefficients.
In the room acquisition, the wrist camera viewed the laboratory
from 25 poses with small relative rotations (at most 24\,deg, and 1.5 to
3.8\,deg between consecutive poses). In the tabletop acquisition, it viewed nearby objects
with more varied rotations.
The fixed camera occupied a different pose in each acquisition.
Its reference poses differ by 98.5\,cm and 178.9\,deg.
All subsets reused these acquisitions, without independent wrist-camera remounts.
Reference extrinsics come from
separate ChArUco~\cite{ArUco} calibrations, using a fixed board for the
wrist camera and an end-effector board for the fixed camera. The RMS
dispersion of the recovered calibration-target poses is 2.8 to
3.8\,mm and 0.56 to 1.56\,deg in the corresponding base or end-effector
frame.
\looseness=-1

\paragraph{Wrist-camera estimation and pair selection}
At $N_w=15$, applying our FK-guided pair selection to the same JCR solver
reduces mean wrist translation error from 10.16 to 3.89\,cm in the room
acquisition and from 2.41 to 1.65\,cm in the tabletop acquisition.
The full pipeline achieves 1.35 and 1.22\,cm, respectively.
In the room acquisition, the small rotations between adjacent poses
provide weak translation constraints for JCR. JCR
and the baseline of Zhi and Schwertfeger exceed 10\,cm at every budget, while our error reaches 1.33\,cm at $N_w=25$
(Section~\ref{sec:exp_components}(b)).

\paragraph{Fixed camera}
Fixed-camera translation errors at $N_w=15$ are 1.63 and 1.75\,cm,
compared with 8.76 and 3.42\,cm for JCR~+~SF. Relative to our
initial fixed-camera estimates (Initial estimate + SF in Table~\ref{tab:wrist_ablation}), the reductions are $3.66$\,cm $[2.16,5.15]$ and
$2.19$\,cm $[1.33,3.19]$, respectively.

\paragraph{Overall comparison}
Across the 24 simulation conditions in Table~\ref{tab:main} 
and the six physical conditions in Table~\ref{tab:physical}, wrist translation and rotation
errors are 80.2\% and 68.1\% lower than JCR's on average, and fixed-camera
errors are 80.9\% and 66.0\% lower than those of JCR~+~SF.

\subsection{Component Analysis}
\label{sec:exp_components}
\label{sec:exp_ext}

\begin{table}[!t]
\caption{Wrist-Camera Refinement Variants and Fixed-Camera Comparisons}
\label{tab:wrist_ablation}
\centering
\scriptsize
\setlength{\tabcolsep}{2.4pt}
\renewcommand{\arraystretch}{0.95}
\begin{minipage}{\linewidth}
\begin{tabular*}{\linewidth}{@{\extracolsep{\fill}}l >{\columncolor{black!6}}r r >{\columncolor{black!6}}r r
  !{\hspace{3pt}} >{\columncolor{black!6}}r r >{\columncolor{black!6}}r r@{}}
\toprule
& \multicolumn{4}{c}{Simulation} & \multicolumn{4}{c}{Physical} \\
\cmidrule(lr){2-5}\cmidrule(l){6-9}
& \multicolumn{2}{c}{Room} & \multicolumn{2}{c}{Tabletop}
  & \multicolumn{2}{c}{Room} & \multicolumn{2}{c}{Tabletop} \\
\cmidrule(lr){2-3}\cmidrule(lr){4-5}\cmidrule(lr){6-7}\cmidrule(l){8-9}
Variant & $e_t$ & $e_\theta$ & $e_t$ & $e_\theta$ & $e_t$ & $e_\theta$ & $e_t$ & $e_\theta$ \\
\midrule
\multicolumn{9}{@{}l}{\textit{Wrist camera}} \\
Initial estimate (no BA) & 2.53 & 0.58 & 1.57 & 0.39 & 3.28 & \textbf{1.32} & 1.48 & 1.30 \\
Exact FK & \textbf{1.02} & \textbf{0.22} & \textbf{0.19} & \textbf{0.10} & 8.06 & 3.71 & \textbf{1.22} & 1.32 \\
FK priors & 2.30 & 0.56 & 0.90 & 0.34 & 2.46 & 2.09 & 1.59 & 1.59 \\
FK priors + lens ref. & 2.49 & 0.61 & 0.98 & 0.33 & \textbf{1.35} & 1.48 & 1.54 & \textbf{0.94} \\
\textbf{Ours} & \textbf{1.02} & \textbf{0.22} & \textbf{0.19} & \textbf{0.10} & \textbf{1.35} & 1.48 & \textbf{1.22} & 1.11 \\
\midrule
\multicolumn{9}{@{}l}{\textit{Fixed camera}} \\
JCR~\cite{JCR} + SF & 5.03 & 0.50 & 8.65 & 0.61 & 8.76 & 3.22 & 3.42 & 2.40 \\
JCR + reg. & 8.15 & 0.39 & 8.11 & 0.59 & 10.12 & 4.04 & 3.58 & 2.66 \\
Initial estimate + SF & 6.44 & 0.78 & 8.80 & 0.66 & 5.28 & 1.84 & 3.94 & 2.25 \\
Initial estimate + reg. & 4.46 & 0.68 & 6.16 & 0.55 & 4.10 & 1.41 & 2.25 & 1.93 \\
Refined wrist + SF & 2.81 & 0.42 & 8.67 & 0.58 & 3.91 & 1.88 & 6.40 & 2.14 \\
\textbf{Ours} & \textbf{1.82} & \textbf{0.28} & \textbf{3.81} & \textbf{0.28} & \textbf{1.63} & \textbf{1.40} & \textbf{1.75} & \textbf{1.46} \\
\bottomrule
\end{tabular*}
\begin{papertablenote}
Mean $e_t$ (cm)/$e_\theta$ (deg), lowest per column and block in bold.
$N_w=15$ wrist poses, VGGT. $M$: number of fixed cameras.
Simulation: ten wrist subsets per scene type common to
all $M$, with fixed-camera results over $M=1$--4. Physical: five
subsets per acquisition.
Ours used exact FK except on all five physical room subsets and two of
five tabletop subsets, which used FK priors with lens refinement.
SF: shared reconstruction frame placement, Eqs.~\eqref{eq:vfm_base_init}--\eqref{eq:external_init}.
Refined wrist + SF adds scale refitting and axial correction,
Eqs.~\eqref{eq:scale_refit}--\eqref{eq:range_correction}.
\end{papertablenote}
\end{minipage}
\end{table}

\paragraph{Scale coupling on real images}
To probe scale coupling without FK error or fixed cameras
(Section~\ref{sec:method_ba}), we used the 256 RealSense views
of GraspNet scene\,0100~\cite{GraspNet}, an object-centered trajectory at
$0.39\pm0.01$\,m.
We constructed virtual end-effector poses by composing the published
camera poses with the inverse of a constant wrist transform. Over ten subsets of 25 views,
mean translation errors for JCR and our initial estimate are 22.31 and
18.25\,cm, almost entirely axial. BA reduces our mean translation error
to 1.08\,cm on this
approximately object-centered trajectory (Fig.~\ref{fig:mechanisms}a),
while the baseline of Zhi and Schwertfeger reaches 1.74\,cm.

\paragraph{Wrist-camera refinement variants}
The wrist-camera variants in Table~\ref{tab:wrist_ablation} share wrist
subsets, initial estimates, and SIFT tracks. Lens parameters remain fixed
in the FK-prior ablation.
Exact FK gives the lowest simulated errors.
At $N_w=15$ in the physical room acquisition,
BA with FK priors and lens refinement reduced the initial estimate's
mean wrist translation error but slightly increased rotation error.
In the tabletop acquisition, the
full pipeline reduced mean wrist translation error from 1.48 to 1.22\,cm,
using relaxation on two of five subsets. The 95\% paired-bootstrap
interval for this reduction included zero.
\looseness=-1

For exact FK, $r_{\mathrm{in}}<0.30$ on every physical room subset.
At $N_w=25$, exact FK is retained on three of five tabletop subsets.

Because the exact-FK and FK-prior variants use different loss and
rejection thresholds, we also evaluate exact FK with both thresholds set
to 4\,px and the lens parameters fixed at $\boldsymbol{\kappa}_0$. Its physical
room mean translation error is 8.09\,cm, versus 2.46\,cm for FK priors and
1.35\,cm with lens refinement. At matched loss/rejection settings, FK
corrections reduced mean translation error in this acquisition but
increased it in both simulated groups and the physical tabletop
acquisition. Lens refinement increased simulated translation errors where
intrinsics are exact.
\looseness=-1

A further BA control fixed the full factory camera model, including all
eight distortion coefficients, using the same tracks, initial estimates,
and 4\,px thresholds as the preceding control.
Replacing exact FK with FK priors reduced room mean
translation error from 6.53 to 1.11\,cm but increased tabletop error
from 0.97 to 1.31\,cm. 
With FK priors, the fixed factory camera model gave lower room mean
translation error than BA with lens refinement initialized at zero
distortion (1.35\,cm). FK priors and lens refinement thus have
acquisition-dependent effects.
\looseness=-1

\paragraph{Stages of fixed-camera calibration}
Over the 360 mixed-camera runs of
Table~\ref{tab:exp_ext_wrist}, mean fixed-camera translation error falls
from 7.28\,cm initially to 5.42\,cm after wrist-camera refinement, scale
refitting, and axial correction, then to 2.61\,cm after registration and
the ambiguity test. Each stage retains all preceding steps.
For each wrist estimate, the fixed-camera block of Table~\ref{tab:wrist_ablation}
uses identical matching, triangulation, PnP, and acceptance rules
(Section~\ref{sec:method_fixed}). All wrist subsets pass the full pipeline's
minimum-support check. Rejected registrations return the fixed-camera
estimate from the corresponding SF row.
Registration using JCR or the initial wrist estimate gives higher
fixed-camera errors than ours in every column. Adding registration to JCR
increased translation error in three of four columns. These comparisons
support refining the wrist transform before registration.

An additional SF control using GT or reference wrist transforms gives
mean fixed-camera translation errors of 3.79 to 8.90\,cm across the four
settings of Table~\ref{tab:wrist_ablation}. Wrist-transform error alone
therefore does not explain the remaining error.

To evaluate axial correction in the fallback estimate, we compared
corrected and uncorrected poses with the refined wrist transform and scale fixed. 
Mean simulated translation error decreased by 5.15\,cm
in room scenes but increased by 1.77\,cm in tabletop scenes, 
including the camera whose registrations were rejected.
\looseness=-1

\paragraph{Correspondences for fixed cameras}
We compared classical SIFT feature matching (4,000 features,
0.80 ratio test) and dense matching on 90 fixed-camera estimates from 20 matched
simulated $M=4$ runs and ten physical $N_w=15$ subsets. The wrist
transform, FK, intrinsics, triangulation, PnP, and acceptance
rule were identical. SIFT and dense matching reached the 30-inlier
minimum for 14 and 90 estimates, respectively, and yielded mean final
translation errors of 4.51 and 1.58\,cm, including every fallback.
On the 14 estimates for which both matchers met the inlier threshold, 
mean translation errors of the first PnP poses, 
before ambiguity testing and fallback selection, were 10.31 and 1.36\,cm.

\paragraph{Registration acceptance}
Over 880 simulated fixed-camera estimates, each weighted equally, the
ambiguity test rejected 55 estimates, all from the left-side camera in one
tabletop scene where repetitive structure supported a wrong first pose.
Across these 880 estimates, mean translation errors were 4.83\,cm for the
corrected shared reconstruction frame pose, 11.32\,cm for registration
without the test, and 2.24\,cm with it. The 825 accepted registrations
averaged 1.29\,cm, compared with 4.05\,cm for the corrected shared
reconstruction frame poses of the same cameras. The 55 rejected
registrations would have averaged 1.62\,m, whereas their fallback estimates
averaged 16.51\,cm.

A sufficiently supported distant pose can cause the test to reject a
correct registration. Acceptance also does not guarantee improvement over
the fallback:
62 of the 440 accepted room registrations increased translation error by
more than 1\,cm relative to the corrected shared reconstruction frame pose.
All 26 attempted physical registrations were accepted.
\looseness=-1

\FloatBarrier
\section{CONCLUSIONS}
\label{sec:conclusion}

AnchorHEC calibrates wrist and fixed cameras in a single scene-based
procedure by using the refined wrist transform and FK to triangulate
metric points for fixed-camera registration. It uses known intrinsics
and requires no targets or robot-body references. 
Simulated errors were similar across reconstruction backbones 
when refinement and registration had sufficient image support. 
Physical validation covered two acquisitions with one
fixed camera each. Calibration after independent wrist-camera remounting
remains untested.
\looseness=-1

Future work will investigate using the estimated camera-to-robot transforms
as camera geometry inputs to vision-language-action policies and evaluate
manipulation under changes in camera placement.
\looseness=-1

\bibliographystyle{IEEEtran}
\bibliography{references}

\end{document}